\documentclass[10pt,twocolumn,letterpaper]{article}
\usepackage{wacv}
\usepackage{times}
\usepackage{epsfig}
\usepackage{graphicx}
\usepackage{amsmath}
\usepackage{amssymb}
\usepackage{url}

\usepackage{enumitem}
\usepackage{subcaption}

\usepackage{cite}
\usepackage{amsfonts}
\usepackage{algorithmic}
\usepackage{textcomp}
\usepackage{xcolor}
\usepackage{dirtytalk}
\usepackage[ruled,vlined]{algorithm2e}
\def\wacvPaperID{845} 

\wacvfinalcopy 

\ifwacvfinal
\def\assignedStartPage{1} 
\fi

\ifwacvfinal
\usepackage[breaklinks=true,bookmarks=false]{hyperref}
\else
\usepackage[pagebackref=true,breaklinks=true,colorlinks,bookmarks=false]{hyperref}
\fi

\ifwacvfinal
\else
\fi

\begin{document}

\title{Deep4Air: A Novel Deep Learning Framework for Airport Airside Surveillance}

\author{Thai Van Phat\\
Saab-NTU Joint Lab\\
Nanyang Technological University, Singapore\\
{\tt\small thai0009@e.ntu.edu.sg}
\and 
Sameer Alam\\
Air Traffic Management Research Institute\\
Nanyang Technological University, Singapore\\
{\tt\small sameeralam@ntu.edu.sg}
\and 
Nimrod Lilith\\
Saab-NTU Joint Lab\\
Nanyang Technological University, Singapore\\
{\tt\small nimrod.lilith@ntu.edu.sg}
\and 
 Phu N. Tran\\
Air Traffic Management Research Institute\\
Nanyang Technological University, Singapore\\
{\tt\small phutran@ntu.edu.sg}
\and 
 Nguyen Thanh Binh\\
Faculty of Mathematics and Computer Science\\
VNUHCM-University of Science, Ho Chi Minh City, Vietnam\\
{\tt\small ngtbinh@hcmus.edu.vn}
}

\maketitle

\begin{abstract}
An airport runway and taxiway (airside) area is a highly dynamic and complex environment featuring interactions between different types of vehicles (speed and dimension), under varying visibility and traffic conditions. Airport ground movements are deemed safety-critical activities, and safe-separation procedures must be maintained by Air Traffic Controllers (ATCs). Large airports with complicated runway-taxiway systems use advanced ground surveillance systems. However, these systems have inherent limitations and a lack of real-time analytics. In this paper, we propose a novel computer-vision based framework, namely \say{Deep4Air}, which can not only augment the ground surveillance systems via the automated visual monitoring of runways and taxiways for aircraft location, but also provide real-time speed and distance analytics for aircraft on runways and taxiways. The proposed framework includes an adaptive deep neural network for efficiently detecting and tracking aircraft. The experimental results show an average precision of detection and tracking of up to 99.8\% on simulated data with validations on surveillance videos from the digital tower at George Bush Intercontinental Airport. The results also demonstrate that \say{Deep4Air} can locate aircraft positions relative to the airport runway and taxiway infrastructure with high accuracy. Furthermore, aircraft speed and separation distance are monitored in real-time, providing enhanced safety management.
\end{abstract}

\section{Introduction}
Major airports worldwide have undertaken substantial expansion programs to accommodate the steady growth in air traffic, including the construction of new runways and taxiways. However, putting new construction into operation increases the challenge of aircraft ground movement control and monitoring. During the last five years alone, approximately 1500 runway incursions have been reported in the US alone, and their frequency has risen annually~\cite{RI:FAA}. A runway incursion can be defined as \say{any occurrence at an aerodrome involving the incorrect presence of an aircraft, a vehicle, or a person on the protected area of a surface designated for landing and take-off of aircraft}~\cite{RI:ICAO}. Consequently, accurately detecting and tracking every (moving) object in the airport airside is vital to reduce runway incursions and maintain situation awareness for ATCs.

Typically, complex airside operations use advanced surveillance systems, such as Advanced Surface Movement Guidance and Control Systems (A-SMGCS)~\cite{ASMGCS}. An A-SMGCS system provides four operational functions: surveillance, control, routing, and guidance. For the surveillance function, the system can use different sensors, including Surface Movement Radar (SMR), Automatic Dependent Surveillance-Broadcast (ADS-B), and Multilateration (MLAT). However, these systems still have drawbacks in terms of accuracy, cooperation, noise, and delay~\cite{caoguidance}. Precisely, ADS-B and MLAT require a transponder to be installed for communication. SMR does not need a transponder, but it can incorrectly detect ground vehicles as aircraft. Moreover, if there are obstacles, such as buildings or metal objects, between the airplane and sensors, the plane's measured position can differ from its actual location~\cite{error}.

According to the International Civil Aviation Organization, controllers continuously watch all flight operations on and in the vicinity of an aerodrome by visual observation, augmented in low visibility conditions by radar when available~\cite{ICAO01}. More precisely, controllers must detect, recognize, and identify the aircraft type, the corresponding airline, and its taxiway design group~\cite{intro:feature, application}. Also, they must control and manage aircraft speed, direction, and location on taxiways~\cite{intro:visual} to minimize any potential risk of collision between different airplanes. With the ambitious growth plans to serve more passengers and make air-travel safe and efficient, airports and air navigation service providers are implementing measures to accommodate technological advances in order to enable ATCs to work efficiently. Thus, the concept of a camera-based surveillance system in an airport has been broadly investigated~\cite{TRAVIS, INTERVUSE, PTZ}. However, these systems exploit traditional machine learning methods to detect and recognize airplanes and ground vehicles in aircraft parking areas.
\begin{figure}
    \centering
    \includegraphics[width=0.42\textwidth]{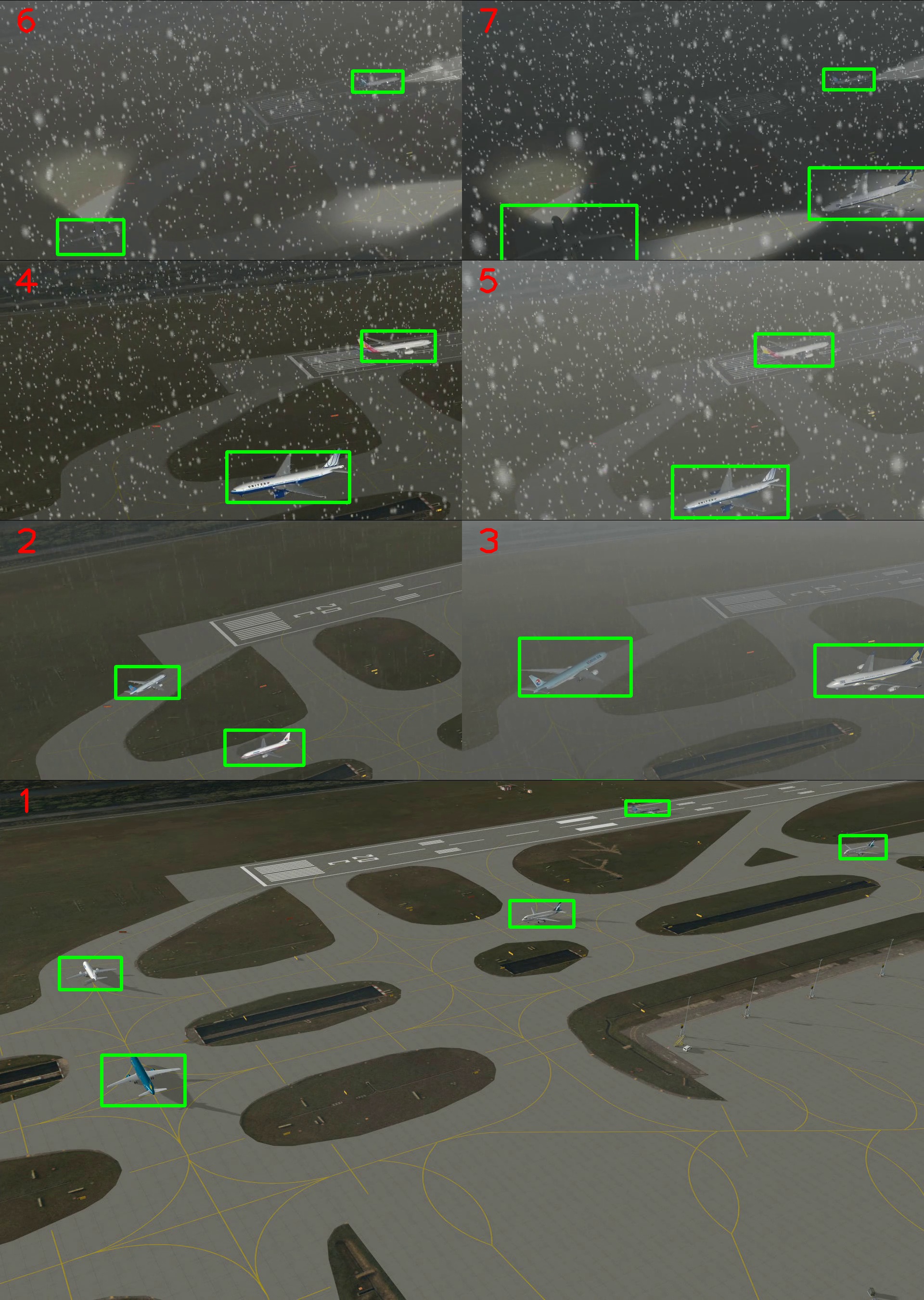}
    \caption{Videos generated by the simulator featuring different environmental conditions. Image (1) is a full frame of a sunny day, while the rest are quarter frames combining rain, fog and snow with dusk and night.}
    \label{fig:sample}
\end{figure}

We propose a novel framework for airside surveillance for airport ground movements to monitor runway and taxiway areas for better airside safety management. The primary objective is to improve safety by exploiting Convolutional Neural Networks (ConvNets) and high-resolution videos to detect aircraft. However, the state-of-the-art ConvNets~\cite{yolov3,rentina, ED} currently require significant processing power to perform real-time detection. To overcome this, we have built a specific ConvNet architecture that can run faster than the state-of-the-art ConvNets, while achieving a similar performance level. After detection, it is possible to track each aircraft in real-time. Furthermore, it can also estimate the aircraft's location, speed, and distance. Finally, by fusing camera and radar information, we can provide aircraft identification, including aircraft type and company. This information can be displayed on the screen with different colors representing different meanings, potentially reducing controller head-down times~\cite{RTO:HDT}. Our framework takes advantage of the fact that digital towers~\cite{RVT,RTO:HDT} feature a network of high-resolution cameras, covering a 360-degree view of an airport, to provide the camera-based digital tower video. Typically, the network has up to 14 cameras, of which 3 to 7 cameras will cover a runway and its vicinity. Besides accuracy, speed is an essential factor in a computer vision surveillance system. By reducing computational time, observation frequency can be increased, potentially leading to enhanced safety.

The main contributions of our work can be summarized as follows. First, we propose a novel framework for airside safety management with the following functions: aircraft detection and tracking in real-time, aircraft localization, and both speed and direction estimation. Second, we present an adaptive ConvNet architecture, which can run faster than the state-of-the-art ConvNets while achieving similar performance. Third, we provide a fusion of camera and radar sources to retrieve the corresponding aircraft type and the company information. Finally, we present a method of providing ATCs with information that assists their tasks without the need to refocus away from the video display.

The rest of this paper is organized as follows. In Section~\ref{set:rel} we briefly review previous related works, and we describe video data in Section~\ref{set:data}. We present our proposed framework as well as construct different ConvNet architectures for aircraft detection and tracking in Section \ref{set:pro}. Then, we detail experiment design and show our experimental results compared to other techniques in Section~\ref{sec:exp}. Section~\ref{set:app} discusses applications for air traffic control operations. Finally, the paper ends with conclusions and future work in Section~\ref{set:con}.

\section{Related Works}\label{set:rel}
\subsection{Object Detection and Tracking}
ConvNets~\cite{AlexNet} have recently demonstrated enormous potential in the computer vision field, including winning the ImageNet Large Scale Visual Recognition Challenge~\cite{imagenet} in 2012. After several refinements, ConvNets~\cite{ResNet, EN} have become the state-of-the-art models in object recognition. It is inevitable to modify ConvNets for different computer vision tasks, including object detection and tracking. In object detection, region-of-interest approaches~\cite{RCNN, faster} have shown their advantages in terms of flexibility and accuracy. However, they tend to run more slowly compared to one-stage approaches~\cite{ssd, yolov3, ED}. We review three of the most successful models based on different strengths, YOLOv3~\cite{yolov3}, RetinaNet~\cite{rentina}, and EfficientDet~\cite{ED}. By unifying several object detection components into a single network, YOLO executes exceptionally rapidly. Moreover, with a new backbone, DarkNet~\cite{yolov3} (which is faster and more accurate compared to ResNet~\cite{ResNet}), YOLOv3 became the fastest ConvNet among the state-of-the-art models at that time. However, without an extra network to narrow down the number of potential objects as two-stage approaches~\cite{RCNN, faster}, one-stage methods suffer from the foreground-background class imbalance problem~\cite{rentina}. By introducing focal loss, RetinaNet can reduce the relative loss for well-classified examples to focus more on hard, misclassified examples. As a result, RetinaNet currently surpasses the performance of all existing two-stage ConvNets. EfficientDet proposes a compound scaling method which uniformly scales different network hyperparameters at the same time. In theory, EfficientDet achieves better accuracy results with less calculation time compared to other models~\cite{ED}.

ConvNets have also been modified for object tracking. These techniques are variously based on classification~\cite{track1}, similarity learning~\cite{track2}, regression~\cite{track3} or correlation~\cite{track4}. Before tracking, an object needs to be located first by a manual or external algorithm. Therefore, another way to track an object is tracking-by-detection. By detecting objects in every frame, the objects can be mapped from current frames to previous frames creating sequences as tracking. As objects in our task do not overlap, due to the height of the digital tower camera as mentioned in Section~\ref{set:data}, we implement tracking-by-detection as it is the simplest, and therefore computationally lightweight, method of tracking.

\subsection{Camera-based systems for airport}
There are different camera-based systems developed for airport airside surveillance in many airports worldwide. The INTERVUSE~\cite{INTERVUSE} system proposes rectangular areas, namely virtual detectors, to indicate the presence of aircraft. A visual sensor data fusion server receives virtual detector signals from every camera. It then sends these signals to a surveillance data server to track objects with additional information from radar. This system's main advantage is the reduction of computational requirements, as only specified areas are processed. However, the system's performance is sensitive to virtual detector configurations and camera calibration and light conditions. The TRAVIS~\cite{TRAVIS} system consists of a scalable network of tracking units that uses cameras to detect moving objects and provide results to a server. The server tracks and visualizes moving objects in the scene and alerts the presence of dangerous situations. The system is flexible in that it can track multiple moving objects and can be deployed in different environments. However, as the system is based on background extraction, it is susceptible to visual occlusion and overlap, and static or slow-moving objects may not be detected. A system using pan-tilt-zoom (PTZ) cameras to detect aircraft in parking zones has been proposed~\cite{PTZ}. Using PTZ cameras and Haar-like feature detection, the system requires neither static view, calibrated cameras, nor moving objects. However, this approach typically suffers from high false-positive detection. Moreover, PTZ cameras only cover a small part of an airport and require a human operator.
Most significantly, all of these systems perform surveillance of the airport apron area and do not provide the airport runway and taxiway areas supervision. The work presented in this paper offers the monitoring of runway and taxiway areas and real-time analytics. To the best of our knowledge, this is the first work using a ConvNet camera-based architecture to provide aircraft monitoring on airport runways and taxiways.
\begin{figure}
    \centering
    \includegraphics[width=0.47\textwidth]{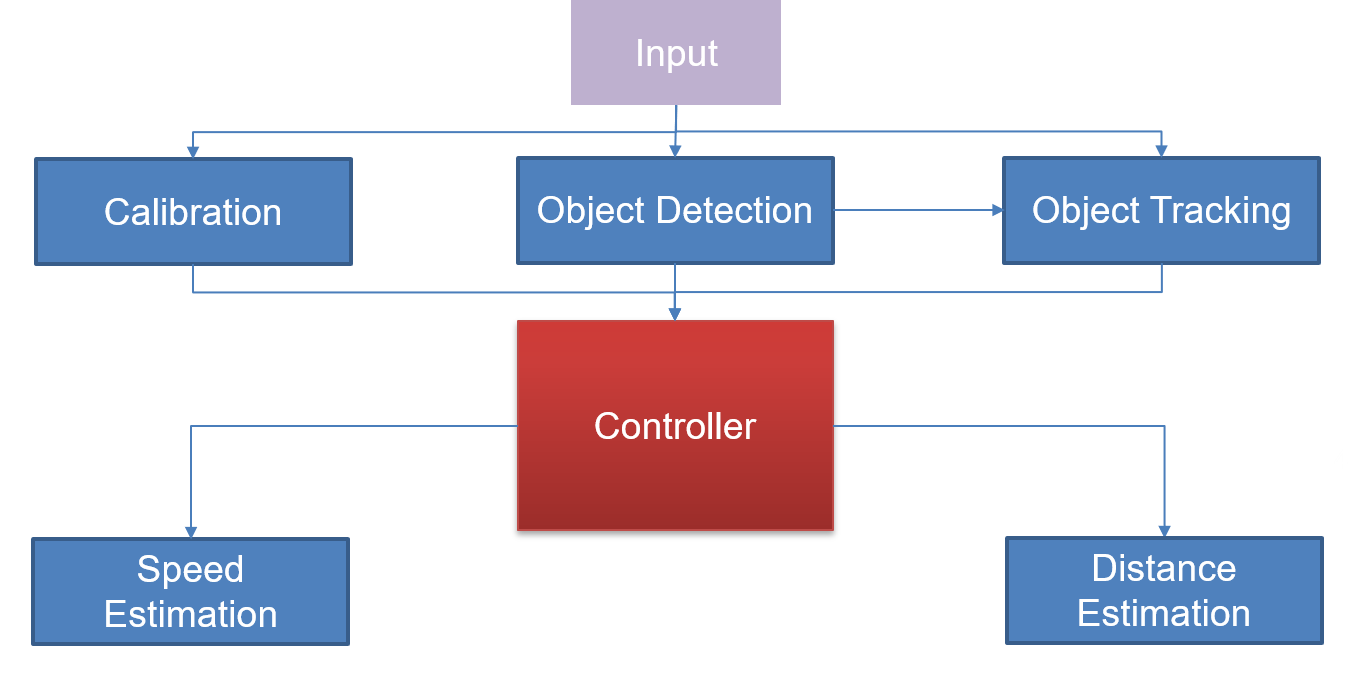}
    \caption{The proposed framework with different functions for controllers. By detecting and tracking every aircraft in runways/taxiways, the system can estimate their speed and distance to notify controllers for any potential risk of runway incursions in an airport.}
    \label{fig:wor}
\end{figure}

\section{Data Collection}\label{set:data}
Due to privacy issues, obtaining sufficient quality videos from digital towers can be extremely difficult. To overcome this, we generated videos of Changi Airport by using the NARSIM simulator system~\cite{narsim} and created two separate sets with the same properties for training and testing. Each set contained videos with seven different visibility conditions, created by combining weather conditions (sunny, rain, snow, and fog), with the time of the day (noon, dusk, and night), as shown in Figure~\ref{fig:sample}. All videos are FHD resolution ($1920 \times 1080$) taken from an $80m$-tall tower. The simulator provides ground truth aircraft information every second, similar to a conventional radar, including \say{call sign}, \say{type}, \say{speed}, and \say{geographic location}. Frames are extracted from the videos and manually labeled as bounding boxes every second to synchronize information between ground truth data and camera. As a result, we create a dataset containing $14184$ training and $14414$ testing images.

Compared to well-known datasets, such as COCO~\cite{coco} and Pascal VOC~\cite{voc}, our video dataset has several differing characteristics. First, our dataset contains high-resolution frames, which will take more time to process. Second, there is only one class, compared to 80 classes from COCO or 20 classes from Pascal VOC. Lastly, as the videos are captured from the tower, aircraft do not overlap each other due to the height of the camera.

It is worth noting that we also managed to obtain real-world videos from cameras at George Bush Intercontinental Airport (IAH), with permission. However, as no corresponding radar ground truth information was included, and the video duration is 30 minutes, they are not suitable for training or validation. However, these videos are used as a demonstration of real-world applicability in our experiments, the details of which can be found in Section~\ref{set:app}.

\section{Methodology}\label{set:pro}
In this section, we present the proposed framework for airside safety management, describe our approaches for aircraft detection and tracking in runways and taxiways, and provide our method for estimating the distances between different aircraft and each aircraft's speed.

\subsection{Overview Framework}
Figure~\ref{fig:wor} shows the framework design, namely \say{Deep4Air} (A \textbf{Deep} Learning Framework \textbf{For} Airport \textbf{Air}side Surveillance), for the problem. Aside from the video input, it also requires the reference location of the airport taxiways, runways, and holding points. Calibration is necessary to map a pixel location to a geographic area. As shown later, object detection is used to detect and track each aircraft in runway and taxiway areas. 

The controller module uses the information from the previous modules for the processing steps, including the assignment of objects to corresponding locations and the estimation of both speed and distance between different airplanes. Speed and distance estimation could then be used to provide advanced warning assistance to ATCs if a safety incident may be likely to occur. This functionality would be of value, as an analysis of the National Transportation Safety Board Aviation Incident Reports~\cite{RI:NTSB} identified a large number of incidents that occurred due to insufficient consideration being given to aircraft separation and/or clear runway status.

\subsection{Input Data}
\begin{figure}
    \centering
    \includegraphics[width=0.48\textwidth]{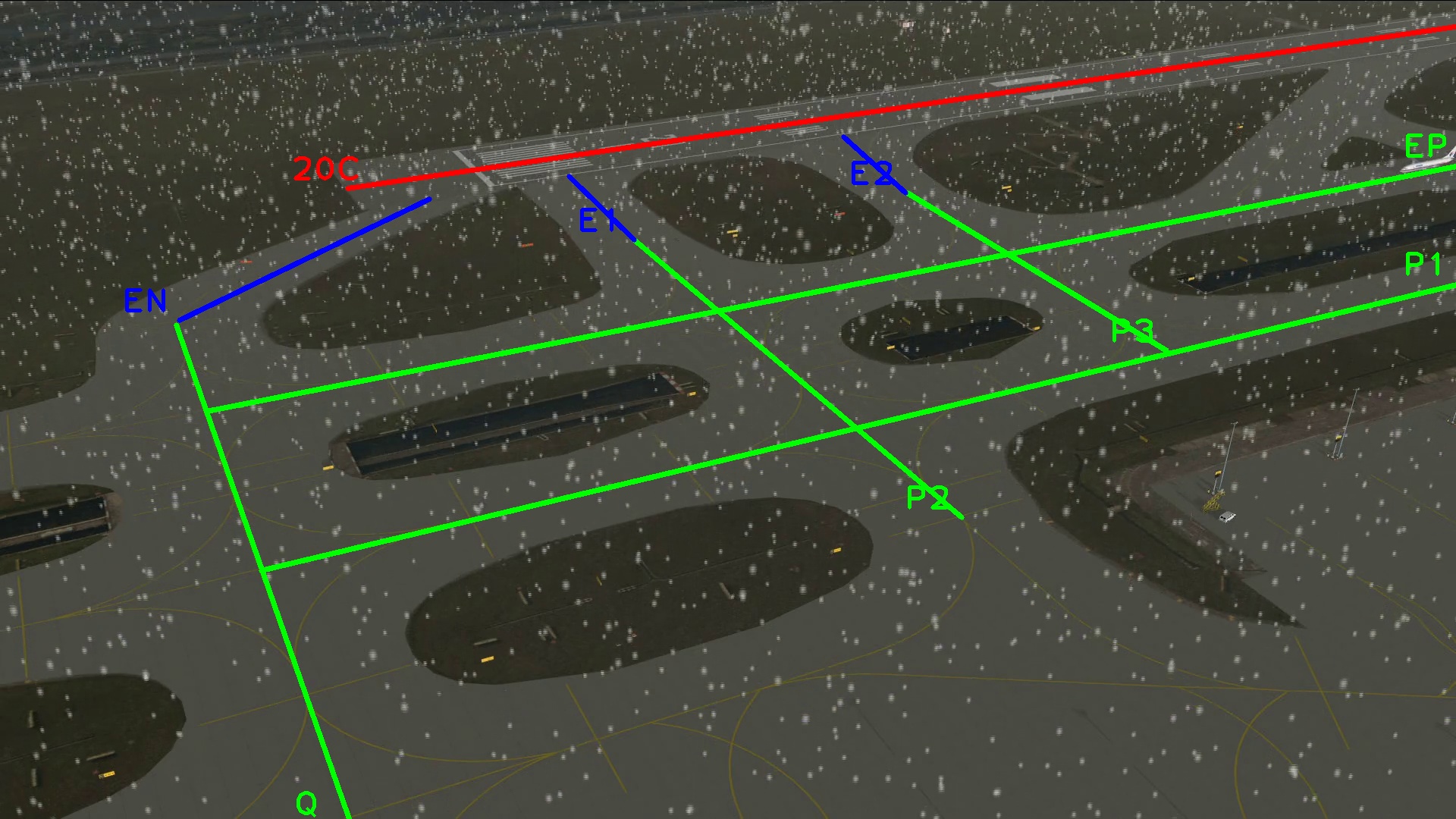}
    \caption{The taxiways (green), runways (red), and holding points (blue) are defined based on their centerlines.}
    \label{fig:ls}
\end{figure}

To assign objects to their corresponding airport locations, we first define those locations. We define them manually, although it is possible to detect them algorithmically. We have taken this approach as the input videos in our experiments are static; therefore, we only need to draw the locations once, as their position relative to the video frames does not change.
Figure~\ref{fig:ls} shows an example of video input. We define three different types of regions, each with two points. The defined region types are taxiways (green), runways (red), and holding points (blue), starting from the holding points to the runway edges. To reduce the search space, we also define the transitional relationships between the different regions. For example, the next legal locations of taxiway EP are taxiway P2, P3, and taxiway Q. By taking this approach, we reduce the total search space for an object, leading to a reduction in computational requirements and increased accuracy. For example, if an aircraft is in taxiway EP, the framework only focuses on taxiway P2, P3, and taxiway Q instead of every region.

\begin{figure*}
    \centering
    \includegraphics[width=0.85\textwidth]{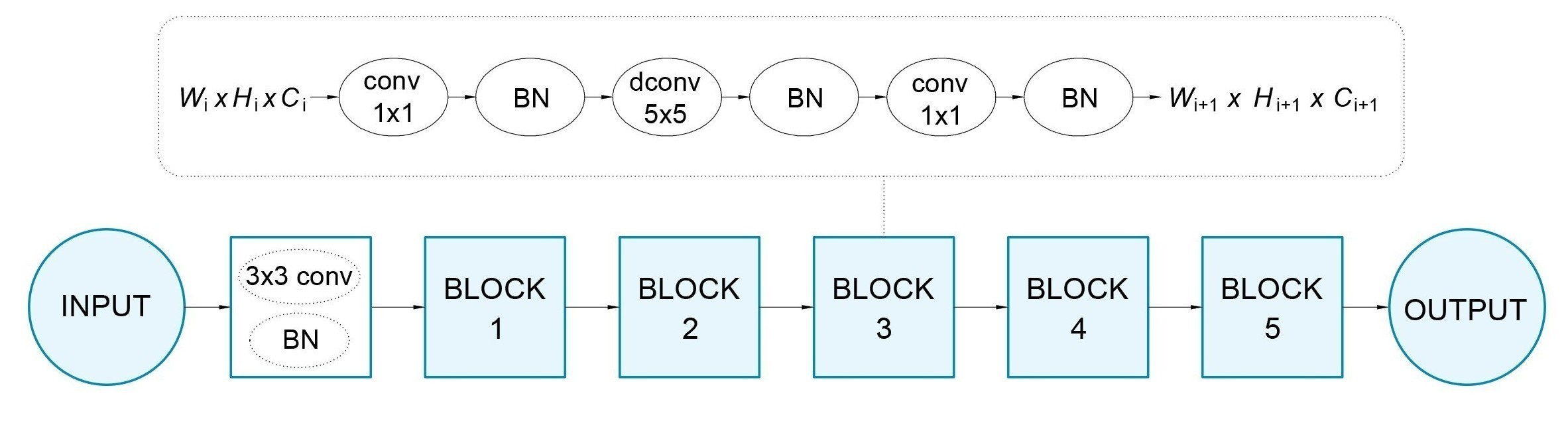}    
    \caption{The ConvNet architecture consists of 5 identical blocks. After each block, the spatial dimensions ($W_i \times H_i $) reduce by half while number of channel ($C_i)$ doubles. The depthwise convolution layers are used in each block.}
    \label{fig:conv}
\end{figure*}

\begin{figure}
    \centering
    \includegraphics[width=0.45\textwidth]{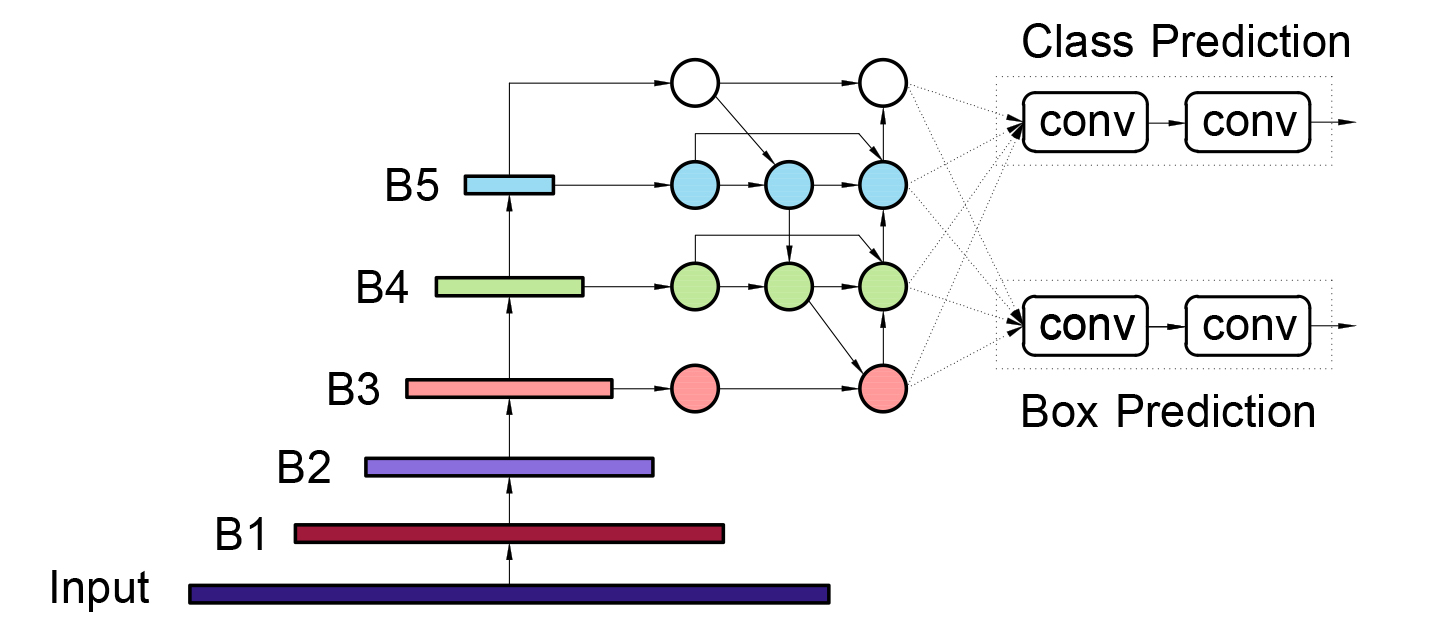}
    \caption{The network architecture. It combines the new backbone with small version of BiFPN to increase its speed.}
    \label{fig:FPN}
\end{figure}

\subsection{Calibration}
To transfer from a video pixel location to a geographic location, we need to calibrate each video. This process is dependent on the camera location of the video source, rather than the framework itself. Since the simulator does not provide camera information, camera calibration is not considered. Instead, we train a machine learning model to translate every pixel point to its corresponding geographic point, using the geographic location of all waypoints provided by the simulator as training labels. Specifically, we build a linear regression model that estimates the geographic location for a given pixel location. We collect $N$ pixel points ($X$) with $N$ corresponding geographic points ($Y$) to construct the data. The linear regression can be written as follows:
\begin{equation}
\tilde{Y} = Wf(X) + \epsilon,
\label{equ:linear}
\end{equation}
where $W$ is the weight learned from data during the training process, $f(.)$ can be viewed as the feature extraction function, and $\epsilon$ is the irreducible error occurred when collecting data. In this work, we choose $f(X)$ as a polynomial function with degree $k$, as shown in Eq.~(\ref{equ:poly}). Here, the chosen degree $k$ should be large enough to prevent underfitting and small enough to prevent overfitting. By experimentation, we choose $k = 5$.
\begin{equation}
    f(x_1, x_2) = \sum_{i=0}^{k} \sum_{j=0}^{i}x_1^jx_2^{i - j}
\label{equ:poly}
\end{equation}

\subsection{Aircraft Detection}\label{set:obj}
Due to the unique video dataset's characteristics, the state-of-the-art ConvNets can easily achieve high accuracy results but do so relatively slowly, as shown in Figure~\ref{fig:ap}. As mentioned above, the speed of detection can enhance safety. Therefore, we create AirNet, a customised ConvNet, which can run faster but still achieve similar performance as the state-of-the-art ConvNets. 

Firstly, the input images are made rectangular, rather than square, with a $W \times H$ ratio of approximately $1.78$. This step is undertaken as the videos are captured from only one source with a resolution of $1920 \times 1080$. By doing this, the computation time is reduced by approximately $10\%$ without compromising detection accuracy.

Secondly, we build a new backbone divided into five identical blocks as shown in Figure~\ref{fig:conv}. The first convolutional layer is used to create $W_0 \times H_0 \times C_0$ from $W \times H \times 3$. Then, the outputs after each block are $W_i \times H_i \times C_i$, where $W_{i+1} = \frac{1}{2}W_i$, $H_{i+1} = \frac{1}{2}H_i$, $C_{i+1} = 1.5C_i$. 
Each block exploits the depthwise convolution, which tremendously reduces the number of parameters~\cite{mobile,EN}. The first $1\times 1$ convolutional layer acts as an expansion layer to uncompress data by expanding the number of channels $W_i \times H_i \times nC_i$. The $5 \times 5$ depthwise convolutional layer with stride 2 aim to filter data and reduce the spatial dimension to half $\frac{1}{2}W_i \times \frac{1}{2}H_i \times nC_i$. The last $1\times 1$ convolutional layer acts as a projection layer to compress data to $\frac{1}{2}W_i \times \frac{1}{2}H_i \times 1.5C_i$. The batch norm layers~\cite{BN} are stacked after the convolutional and depthwise convolutional layers. 

Thirdly, we modify a bidirectional feature pyramid network (BiFPN)~\cite{ED} as a featured network. Since we prioritize speed, our feature network can be viewed as a small implementation of BiFPN, as shown in Figure~\ref{fig:FPN}.

\subsection{Aircraft Tracking}
\begin{figure}
    \centering
    \includegraphics[width=0.45\textwidth]{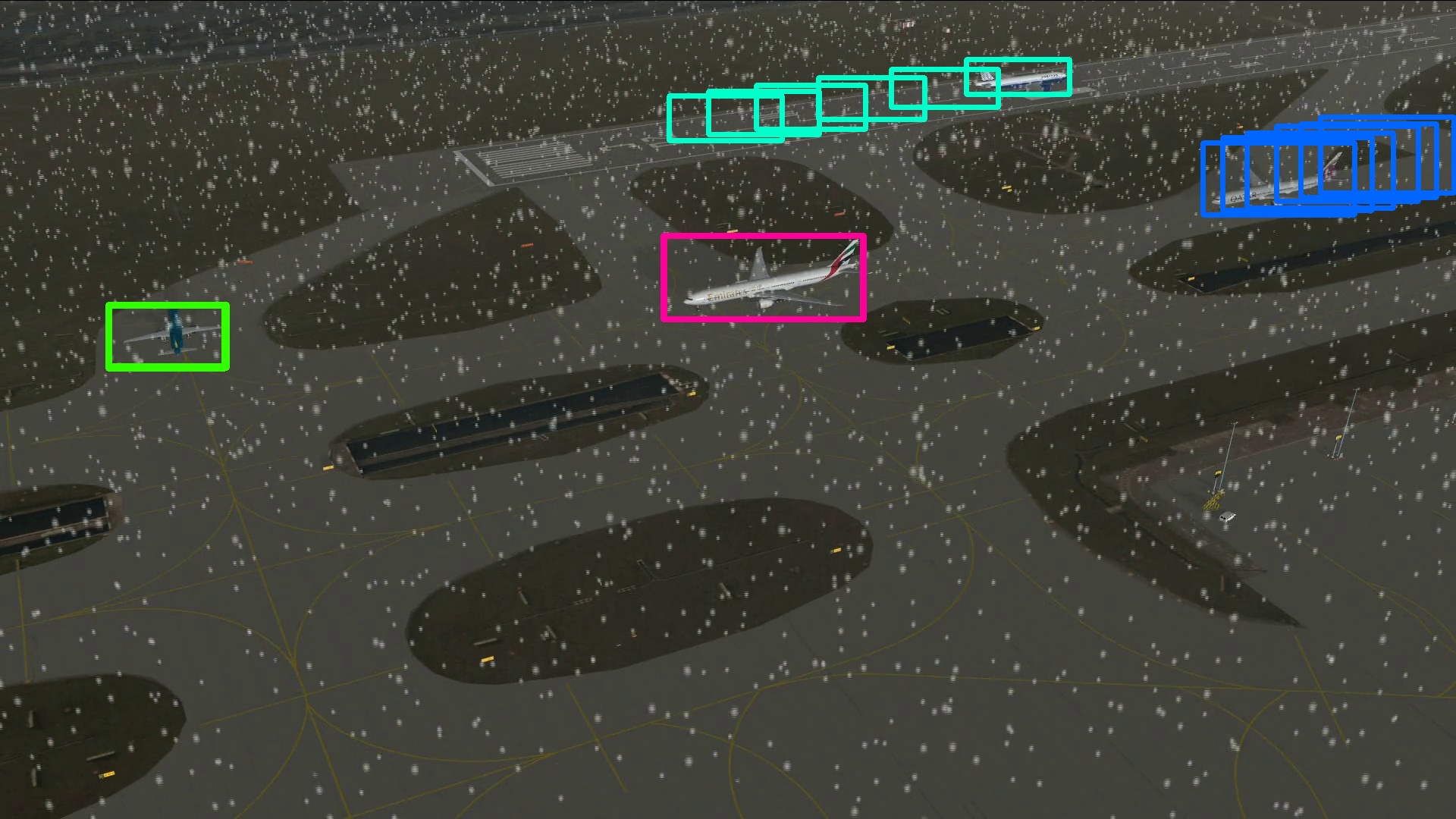}
\caption{Aircraft tracking by detection. By comparing aircraft bounding boxes from previous frames, the algorithm can map them with the current frame.}
    \label{fig:detect}
\end{figure}

We apply the aircraft detection algorithm, as explained above, to detect aircraft. For comparison purposes, we monitor all aircraft at the same rate as the radar (every second). By comparing each bounding box of the current frame with the previous frame's bounding boxes, we can map the aircraft with last movement sequences. Figure~\ref{fig:detect} shows an example of an aircraft detected and tracked by the proposed detection algorithm. 


\subsection{Region Assignment}
\begin{figure}
    \centering
    \includegraphics[width=0.45\textwidth]{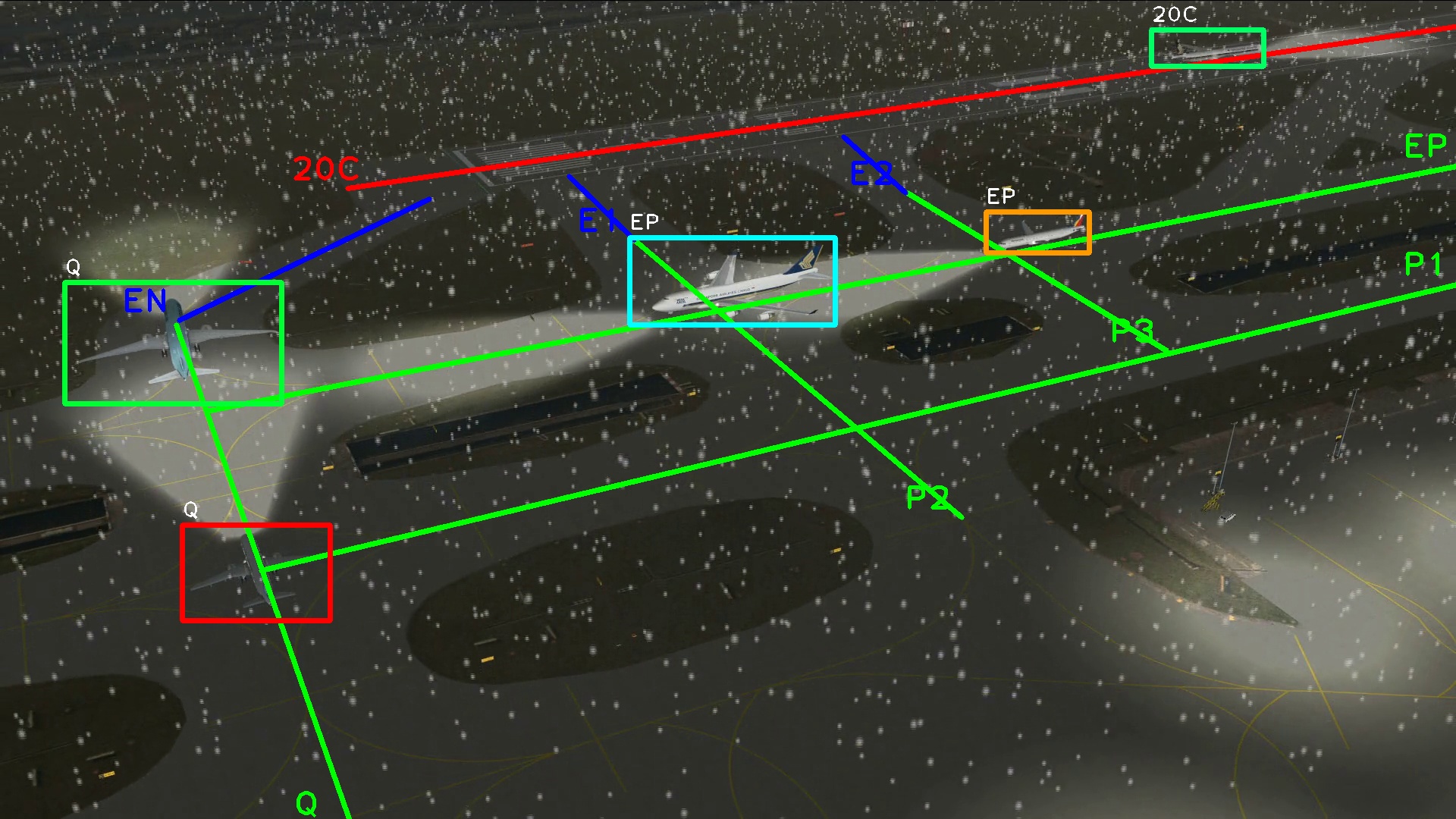}
    \caption{The region assignment function assigns aircraft to specific regions.}
    \label{fig:assig}
\end{figure}

The assignment process is described as follows. First, we calculate intersections between each aircraft and the regions. If there is no intersection, the aircraft does not belong to any region. If there is only one intersection, we assign the aircraft to that region. If there is more than one intersected region, we need to calculate the aircraft's speed. If the aircraft is stationary, the aircraft is assigned to the same region as the previous time. If it is moving, we calculate its direction using its tracked trajectory, ranging from $0^0$ to $360^0$, and compare it with the region directions. The aircraft is mapped to the region having the smallest difference in direction. Figure~\ref{fig:assig} shows the region assignment results.



\subsection{Speed and Distance Estimation}
We can estimate aircraft speed by calculating the distance of a sequence of points over time. Intuitively, we choose the center points of bounding boxes resulting sequence of points $X = {(x_1^{t}, x_2^{t})}, t\in{(1, ...,T)}$. Considering the bounding boxes tend not to be stable, a moving average with step $n$ is applied as $\Bar{x}_i^{(t)} = \frac{1}{n} \sum_{j = t - n}^{t}x_i^{(j)}$.
Then, $\Bar{X}$ are transferred to geographic locations $Y$ by Eq.~(\ref{equ:linear}) with $\epsilon = 0$. Finally, the speed is calculated by the distances divided by the times between two consecutive points. Also, we can compute the corresponding distance by the haversine formula~\cite{caldist}:

\begin{equation}
\begin{split}
dy_i & = y_i^{(t)} - y_i^{(t - 1)}\\
a^{(t)} & = \sqrt{\sin^2{\frac{dy_1}{2}} +\cos{y_1^{(t)}} \cos{y_1^{(t- 1)}} \sin^2{\frac{dy_2}{2}}}\\
d^{(t)} & = 2R \arcsin{a^{(t)}}; R = Average Earth Radius
\end{split}
\label{equ:dist}
\end{equation}

\begin{figure}
    \centering
    \includegraphics[width=0.45\textwidth]{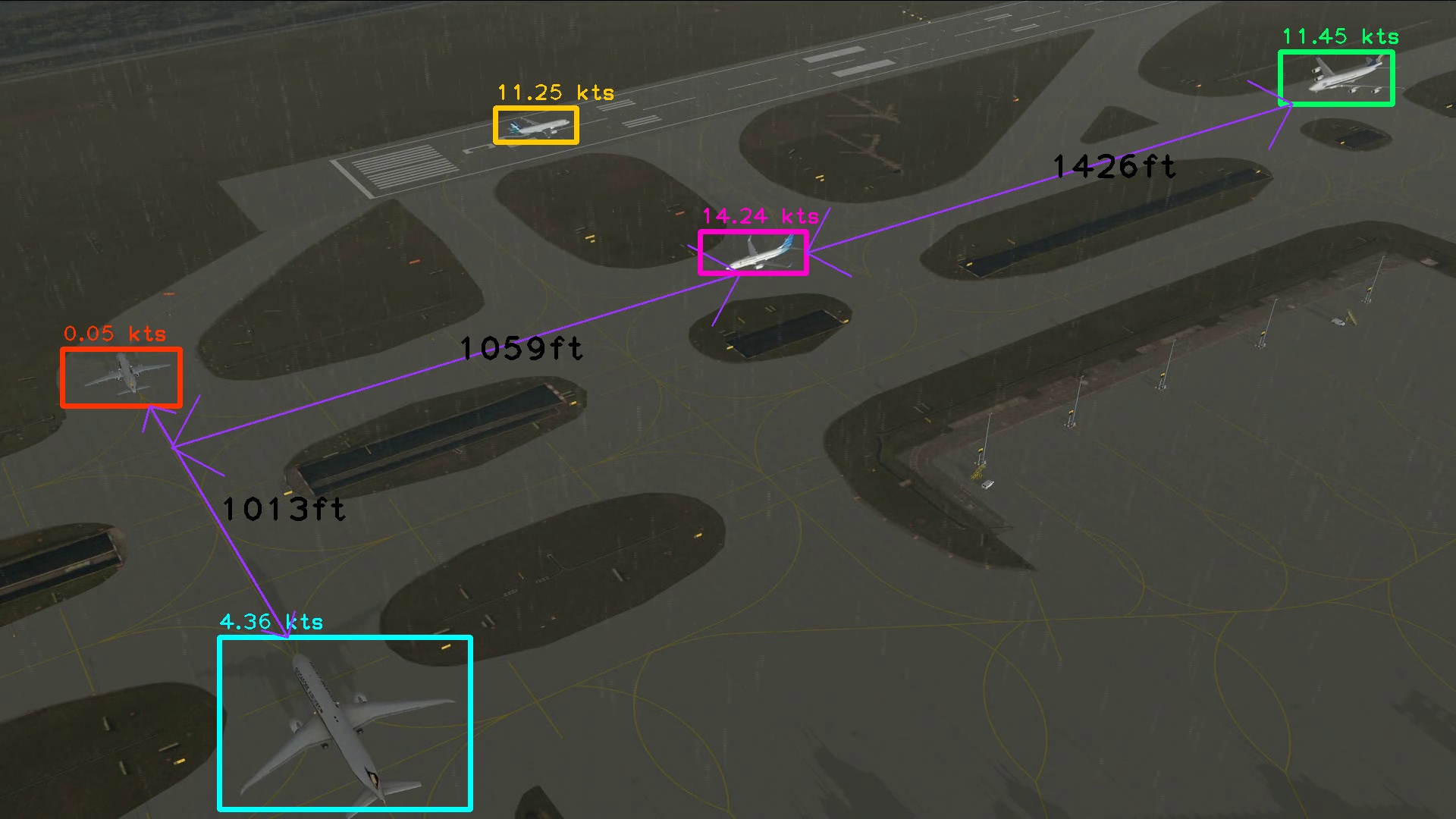}
    \caption{Aircraft speed and the distance between two aircraft or between the aircraft to the next region are estimated.}
    \label{fig:est}
\end{figure}

The distances between each pair of airplanes in the same region are estimated. First, we need to find every pair of aircraft in the same area. Next, intersection points between each plane and the region are computed. Depending on aircraft position, there are one or two intersection points for each aircraft. These two points represent the head and the tail of the plane. Therefore, we can calculate the following four distances between each pair of airplanes: from the tail to the tail, the head to the head, the head to the tail, and the tail to the head, by Eq.~(\ref{equ:dist}). The shortest distance is from the tail of the leading aircraft to the head of the trailing aircraft.
The distances from aircraft to the next regions are also estimated. Similar to the distance between two planes, intersection points between the aircraft and the current area can be calculated, representing the aircraft head and tail. The point in the next region is estimated as an intersection between that region and the current region by Eq.~(\ref{equ:dist}).
Figure~\ref{fig:est} shows the estimated results. The speed units are knots, while distance units are feet.

\section{Experiments and Results}\label{sec:exp}
\subsection{Training ConvNets}

The training image set is split into training and validation sets in the training process, with a ratio of 9:1, respectively. Therefore, we have 12766, 1418, and 14414 images on the training, validation, and test sets. We choose YOLOv3 with Darknet53, RetinaNet with Resnet50, and EfficientDet with EfficientNet to compare with AirNet. First, the AirNet backbone is trained while the rest use backbones trained by ImageNet~\cite{imagenet}. Then, we train four models with the same scheme. We use Adam optimizer~\cite{adam} with learning rate $1e-3$. We implement a learning schedule and an early stop mechanism for the experiments. 
For the learning schedule, we reduce the learning rate by a factor of ten once validation loss no longer decreases in a period of three epochs. If validation loss does not decrease in a period of ten epochs, the training process is stopped.
Hyper-parameters are tuned for speed/performance trade-off. In YOLOv3 and RetinaNet, the only hyper-parameter is the input resolution, while AirNet and EfficientNet also have several layers and filters.
We operate test experiments on a computer with two Intel(R) Xeon(R) Silver 4114 CPUs with 40 threads running at 2.2GHz, 32GB of RAM, and an Nvidia GeForce GTX 1080 GPU. We calculate speed on each image in the test set and report the average time. 
\begin{figure}
    \centering
    \includegraphics[width=0.42\textwidth]{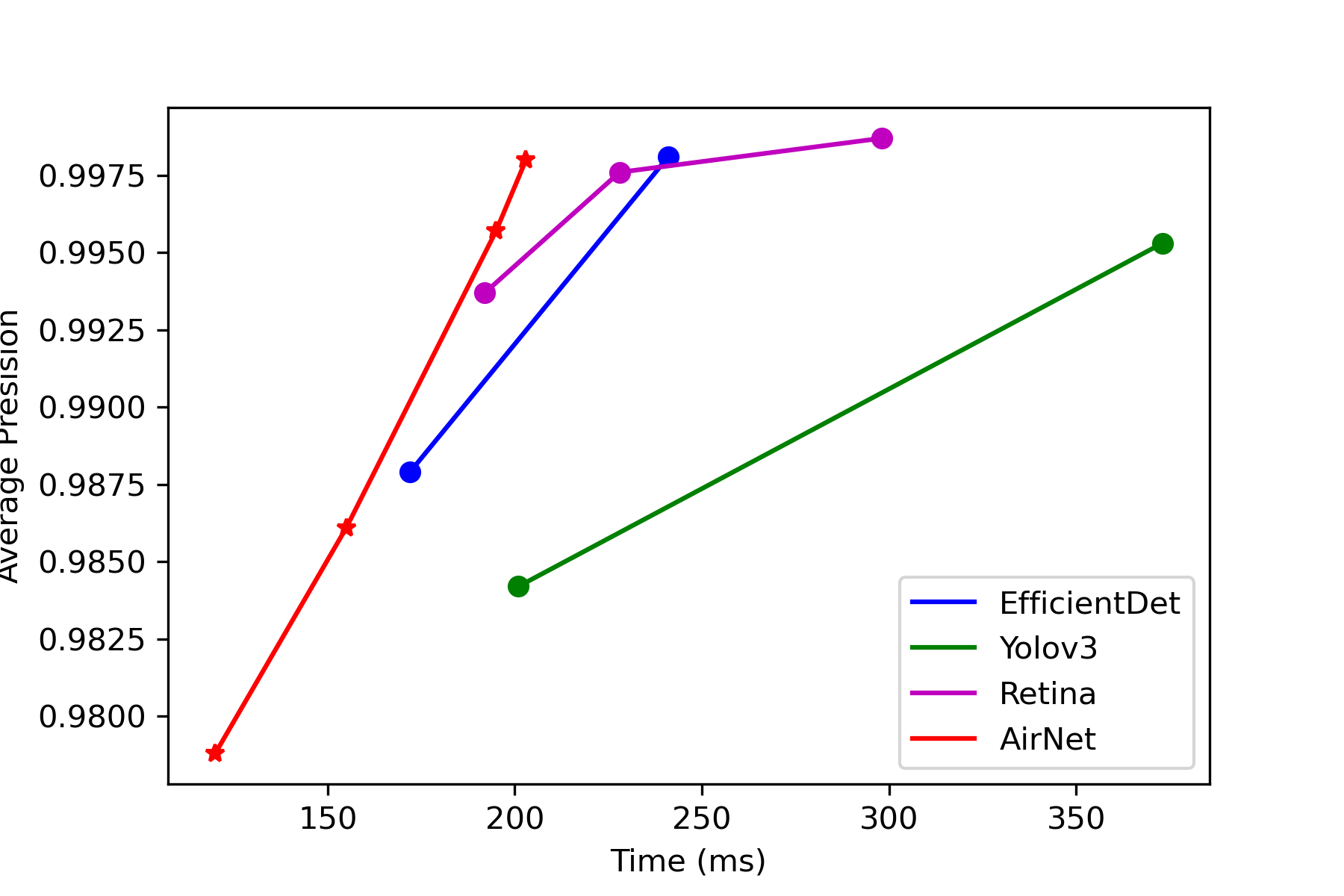}    
    \caption{The model running time versus the corresponding performance on simulated dataset in our experiments.}
    \label{fig:ap}
\end{figure}

\begin{table}
    \begin{center}
        \begin{tabular}{ |c|c|c|c|c|c| }
        \hline
        Model & Res & Params & Time & AP \\ 
        \hline
        Retina & 896 & 36M & 192 & 0.9937\\
                & 1408 & 36M & 228 & 0.9976\\
                & 1920 & 36M & 298 & 0.9987\\
        \hline
        YOLOv3  & 896 & 61M & 201 & 0.9842\\
              & 1408 & 61M & 373 & 0.9953\\
        \hline
        EfficientDet & 1152 & 4M & 172 & 0.9879\\
                    & 1408 & 7M & 241 & 0.9981\\
        \hline
        AirNet & 1152 & 0.25M & 127 & 0.9788\\
                & 1408 & 0.32M & 155 & 0.9861\\
                & 1664 & 0.58M & 195 & 0.9957\\
                & 1920 & 1.03M & 230 & 0.9980\\
        \hline
        \end{tabular}
    \end{center}
    \caption{The experimental results, including the input resolution, the number of parameters, the running time and the average precision, from different networks.}
    \label{tab:AP}
\end{table}

\subsection{Results}
\subsubsection{Aircraft Detection}
Table~\ref{tab:AP} and Figure~\ref{fig:ap} presents the comparison of these ConvNets. As there is only one class and no overlap, all ConvNets can achieve high results. Since YOLOv3 uses Darknet53 with a huge number of parameters, the execution time is slower compared to the other architectures. Importantly, our model runs faster, whilst still achieving high performance, when compared to other models. Moreover, AirNet has fewer parameters, which are more easily loaded on low-power machines. 

\begin{figure}
    \centering
    \includegraphics[width=0.42\textwidth]{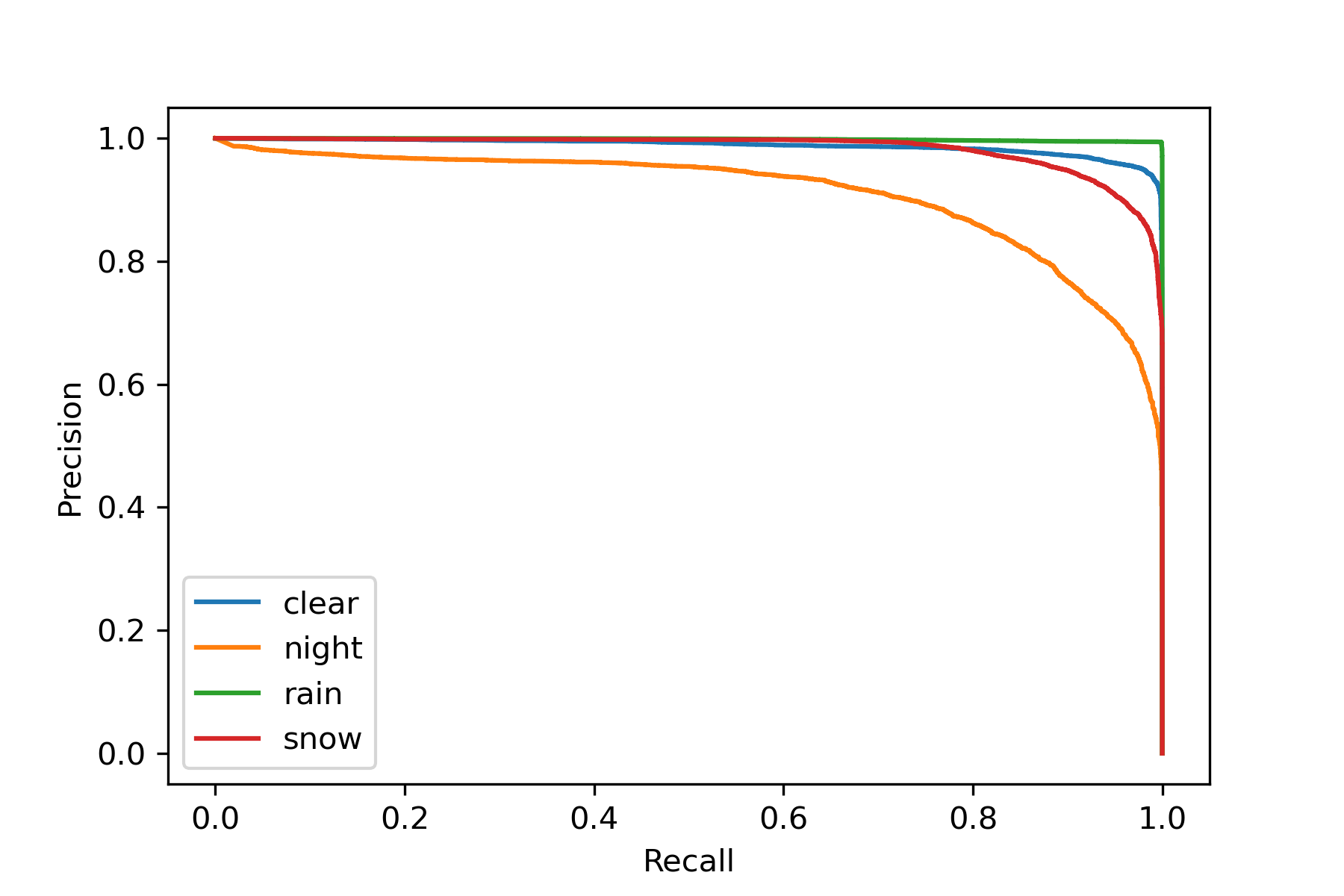}    
    \caption{A precision-recall curve of AirNet@1152 with different conditions.}
    \label{fig:curve}
\end{figure}

In tracking-by-detection, false-negative detecting in some frames is acceptable as objects still can be mapped of the last detected frames. Therefore, we choose AirNet@1152 for the framework detector. Figure~\ref{fig:curve} shows the precision-recall curve of the model with different visibility conditions. As expected, object detection at night has the lowest performance. Also, detecting an object during rain is easier than during snow. 

\begin{figure}
    \centering
    \includegraphics[width=0.42\textwidth]{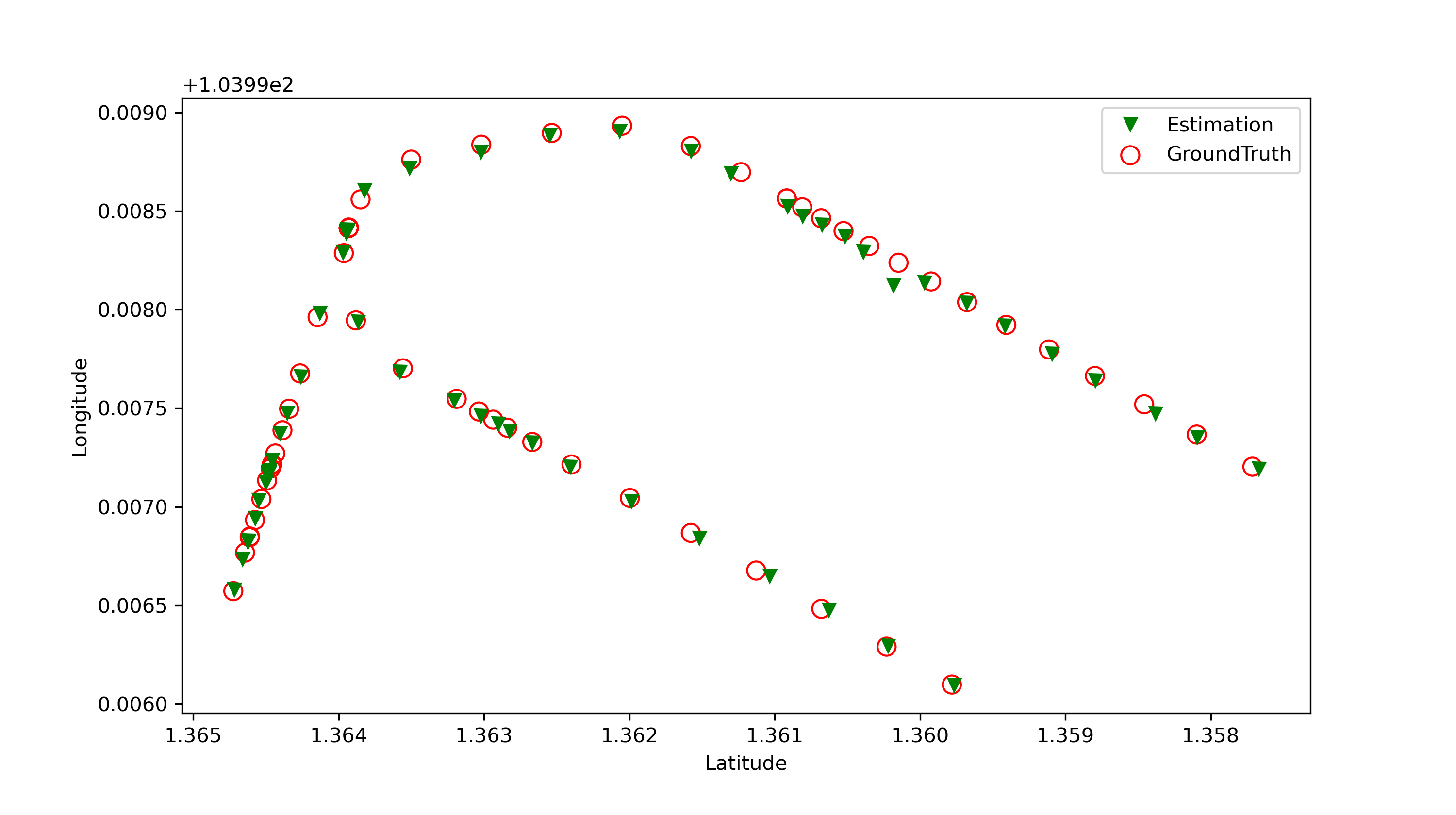}
    \caption{Estimated and actual aircraft positions chosen randomly.}
    \label{fig:point}
\end{figure}

\subsubsection{Speed and Distance Estimation}
Speed and distance estimation are calculated using geographic coordinates, as shown in Eq.~(\ref{equ:dist}). It follows that if the aircraft's estimated position error is minimized, then the speed/distance estimation error will also be minimized. The aircraft's position error of the ASDE-X system~\cite{ASDE} has been measured with a mean of $6.99m$, with individual position errors sometimes exceeding $24.1m$~\cite{error}. By validating $90$ aircraft movements with $21524$ points, the proposed framework's average position error achieves similar performance, with a mean of $7.09m$. The position error percentiles can be visualized in Table~\ref{tab:error}. Figure~\ref{fig:point} shows estimated and actual aircraft positions chosen randomly with mean errors of $5.98m$.
\begin{table}
    \begin{center}
        \begin{tabular}{ |c|c|c|c|c|c| }
        \hline
        Percentiles($\%$) & 5 & 25 & 50 & 75 & 95 \\ 
        \hline
        Error (m) & 1.13 & 3.29 & 5.85 & 9.59 & 16.88\\
        \hline
        \end{tabular}
    \end{center}
    \caption{Different percentiles of aircraft position error.}
    \label{tab:error}
\end{table}

As the speed and distance estimation errors can theoretically be twice the position errors over two successive frames, we therefore implement a moving average speed estimation scheme.

\section{Validation and Applications}\label{set:app}
A list of visual features has been identified for tower controllers~\cite{application}. In this section, we examine the provision of these functions.

\begin{figure}
    \centering
    \includegraphics[width=0.45\textwidth]{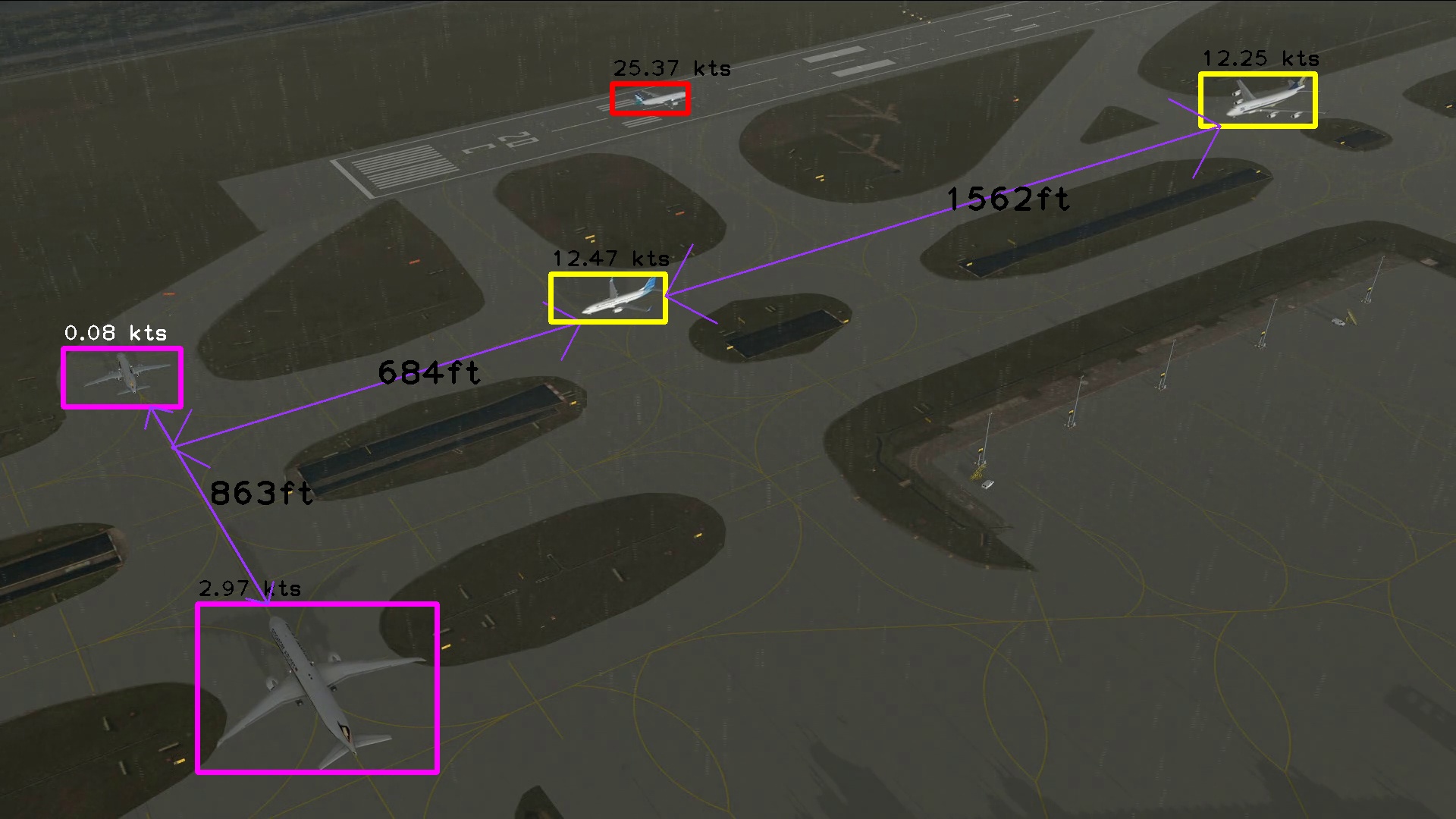}
    \caption{Aircraft information, including moving or stationary aircraft and their location, is displayed by colors.}
    \label{fig:app1}
\end{figure}

First, when controllers look at the screen, it is vital to ensure that they can obtain information as soon as possible. The earlier safety deviations are detected by ATCs, the more time they have to deal with them proactively~\cite{RTO:Safety}. By using colors, the framework can indicate whether aircraft are moving or stationary, or their current regional location. Figure~\ref{fig:app1} shows an example of displaying these types of information. Aircraft on the same regional locations have the same colors. Also, black text indicates moving aircraft while white text indicates stationary aircraft, which can assist in the prediction of separation violations.



\begin{figure}
    \centering
    \includegraphics[width=0.45\textwidth]{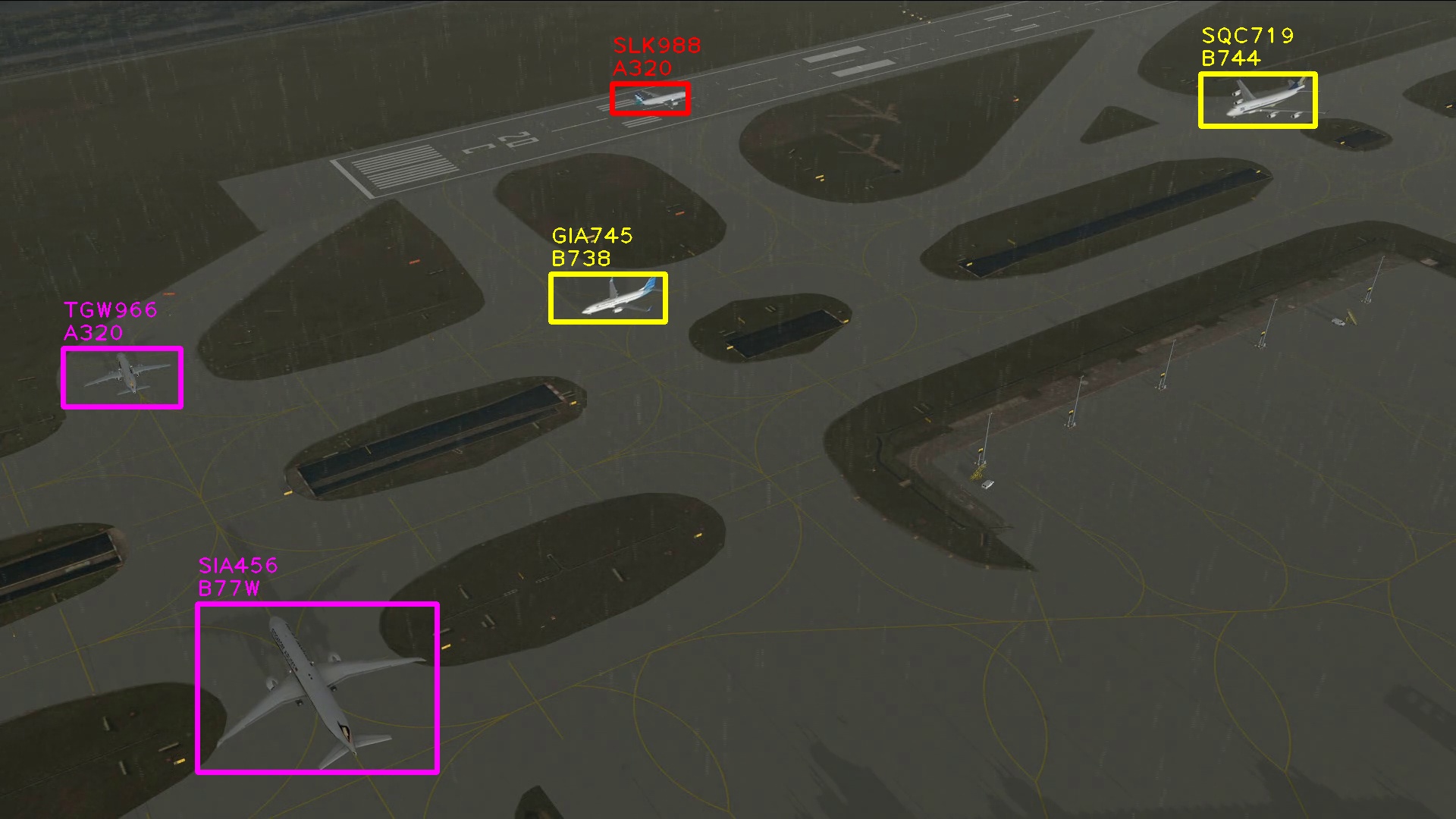}
    \caption{Aircraft information from radar including callsign and type is displayed on the screen.}
    \label{fig:app2}
\end{figure}

Second, after aircraft detection and recognition, the controllers need detailed information such as aircraft callsign and types. More aircraft information can be retrieved from the aircraft type, such as the length, wingspan, the number of engines, and so on~\cite{In:FAA}. Currently, this type of information is challenging to obtain via camera. However, by the fusion of data from camera and radar sources, the system can display this information via video overlay, as shown in Figure~\ref{fig:app2}. By tracking every aircraft, we know their positions as pixel coordinates over time. By calibration, we can translate these pixel coordinates to a sequence of geographic locations over time. Finally, we can map aircraft on the screen to the aircraft radar tracks by comparing these geographic locations.

\begin{figure}
    \centering
    \includegraphics[width=0.45\textwidth]{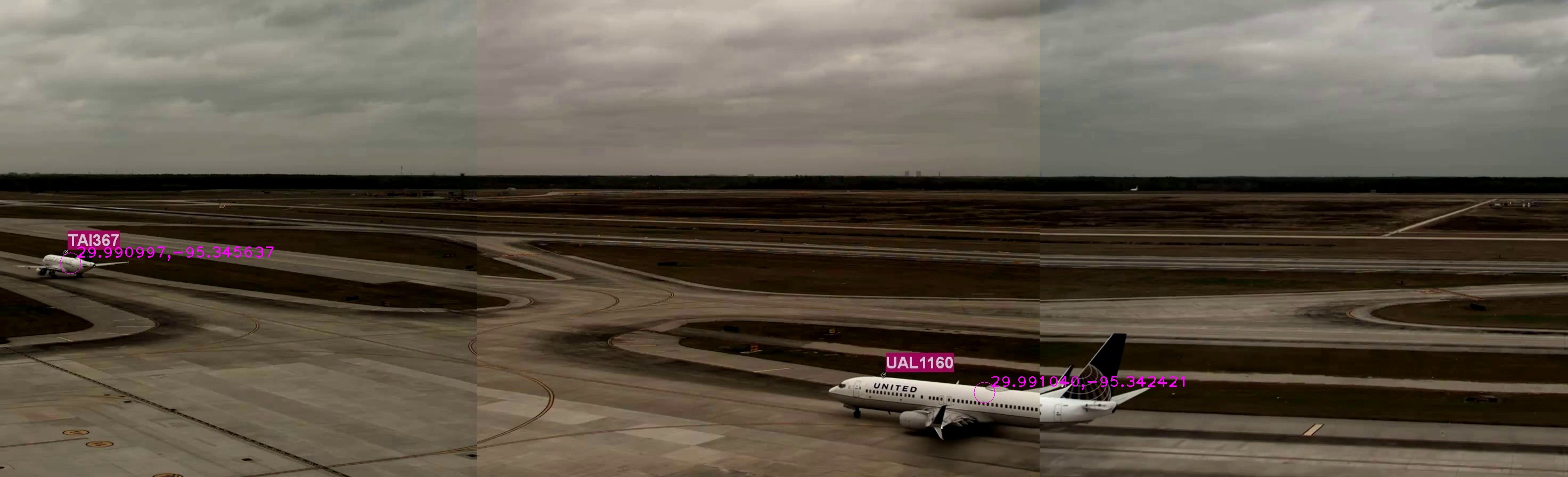}
    \caption{George Bush Intercontinental Airport captured by three cameras from the digital tower with geographic coordinate provided by Deep4Air framework.}
    \label{fig:houston}
\end{figure}

\begin{figure}
    \centering
    \includegraphics[width=0.45\textwidth]{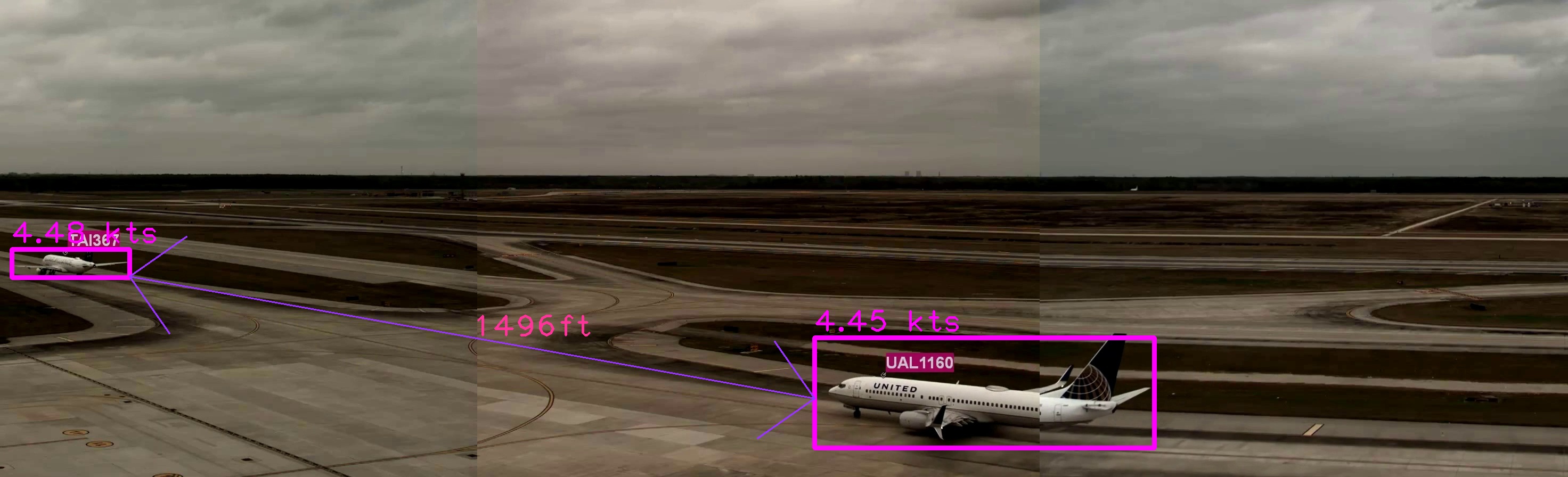}
    \caption{Aircraft detection and speed/distance estimation by Deep4Air framework.}
    \label{fig:houston1}
\end{figure}

As a demonstration of the Deep4Air framework, we use real-world videos from a digital tower at George Bush Intercontinental Airport. There are 14 FHD cameras producing eight frames per second (fps). As our detector configuration takes 127ms for a $1920 \times 1080$ image, we need to pre-process the videos to meet the real-time requirement. First, we select three cameras that capture the main part of the runway and taxiways. Next, the sky background is removed to reduce computation resulting in a $2944 \times 896$ image. The framework not only provides geographic coordinates, as shown in Figure~\ref{fig:houston}, but also detects aircraft and estimates speed and distance at a rate of 6 fps, as depicted in Figure~\ref{fig:houston1}.

\section{Conclusion}\label{set:con}
We have proposed a novel framework that can monitor airport runways and taxiways, and perform essential tasks including assigning aircraft to corresponding regions, estimating aircraft speed, the distance between two aircraft, and the distance of aircraft to their next areas. We also presented an efficient deep learning model for aircraft detection with high average precision (99.8\%) and promising experimental results. As the framework features a high update rate and can detect and track non-cooperative entities, it overcomes these radar surveillance system limitations. With these results, the Deep4Air framework can be used as the primary system for small or medium airports, or a secondary surveillance system for larger airports. 

We intend to extend this work by integrating the framework with real-world videos and conducting human-in-the-loop trials. Also, by combining actual flight plan data with distance and speed estimation, we can investigate collision prediction functionality.
{\small
\bibliographystyle{ieee_fullname}
\bibliography{egbib}
}
\end{document}